# GmFace: A Mathematical Model for Face Image Representation Using Multi-Gaussian

Liping Zhang, Weijun Li, Lina Yu, Xiaoli Dong, Linjun Sun, Xin Ning, Jian Xu, and Hong Qin

*Abstract*—Establishing mathematical models is a ubiquitous and effective method to understand the objective world. Due to complex physiological structures and dynamic behaviors, mathematical representation of the human face is an especially challenging task. A mathematical model for face image representation called GmFace is proposed in the form of a multi-Gaussian function in this paper. The model utilizes the advantages of two-dimensional Gaussian function which provides a symmetric bell surface with a shape that can be controlled by parameters. The GmNet is then designed using Gaussian functions as neurons, with parameters that correspond to each of the parameters of GmFace in order to transform the problem of GmFace parameter solving into a network optimization problem of GmNet. The face modeling process can be described by the following steps: (1) GmNet initialization; (2) feeding GmNet with face image(s); (3) training GmNet until convergence; (4) drawing out the parameters of GmNet (as the same as GmFace); (5) recording the face model GmFace. Furthermore, using GmFace, several face image transformation operations can be realized mathematically through simple parameter computation.

## I. Introduction

AS a visual external biological feature, the face is an important channel for human to convey rich information.

Identity, expression, emotions, intentions, gender, age, ethnic background, attractiveness, and numerous other attributes can be derived from the face of an individual [1]. Over the past decades, significant advances have been made by scholars in face recognition, expression synthesis, and facial animation [2], [3], [4], [5], and the field of face perception has firmly established itself as a significant and active sector within vision research.

When reviewing studies on facial images, existing approaches are commonly based on feature analysis, which can be classified into two main categories: traditional methods based on hand-crafted features [3], [4], [5], [6] , [7], [8] and data-driven deep neural network learning methods [9], [10], [11], [12], [13], [14]. In the traditional methods category, face images are described by a number of digital features extracted from the local or holistic region. Numerous representative methods based on local features have emerged, including local binary pattern (LBP) [3], Gabor wavelet kernel [4], scale invariant feature transform (SIFT) [5], and histogram of oriented gradient (HOG) [6]. Techniques that focus on holistic features, including principal component analysis (PCA) [7] and linear discriminant analysis (LDA) [8] have also emerged. Its basic idea is to get the feature vector from high-dimensional pixel space to subspace by solving the defined objective function optimally. These traditional methods are widely used in texture classification, face detection and recognition. For data-driven methods, researcher use an end-to-end learning mode to obtain a feature set based on face image data and target task training [9]. In recent years, motivated by the classical convolution networks，such as LeNet[16], AlexNet [15], VGG-Net [17], GoogleNet [18], and ResNet [19]，which have achieved excellent results on ImageNet, a number of typical network architectures such as DeepFace [12], FaceNet [13], and SphereFace [14] have been designed for face feature representation.

The methods mentioned above have been demonstrated to successfully obtain the identity features of human faces. Using these advanced methods, some tasks related to human face images can be accomplished and many interesting and valuable applications can be developed. However, whether the face image is transformed from high-dimensional pixel space to encoding feature space manually or treated as a "black box" system [20], [21], [22] to accomplish the task goal automatically, there is still little known about the face structure itself. It remains difficult or potentially impossible to provide a simplified representation to describe human face.

As the most familiar body part, the face shares the common morphological features of eyes, a nose, and a mouth. However, the face is also an aggregation with complex structure in which all facial organs are distinct and can be assembled differently. Even if the topological structure of each face is similar, individual identity can still be distinguished by facial features. Furthermore, affected by age, environment, psychology, and other factors, even face images from the same person can display a rich variety of appearances at different times,

L. Zhang, W. Li, L. Yu, X. Dong, L. Sun, X. Ning, J. Xu, and H. Qin are with the Institute of Semiconductors, Chinese Academy of Sciences, Beijing, 100083, China; Center of Materials Science and Optoelectronics Engineering & School of Microelectronics, University of Chinese Academy of Sciences, Beijing 100049, China; Beijing Key Laboratory of Semiconductor Neural Network Intelligent Sensing and Computing Technology, Beijing 100083, China. Email: {zliping, wjli, yulina, dongxiaoli, sunlinjun, ningxin, xujian, qinh}@semi.ac.cn.

including a smile, sadness, anger, disgust, surprise, and fear [23]. Due to such differences and dynamics, mathematical representation of the human face is an especially challenging task. Establishing mathematical models has remained a significant method to understand the objective world. Based on this concept, despite numerous difficulties, this paper attempts to approximate the mathematical representation of the human face through mathematical modeling.

Intuitively, as the face and its organs exist in the form of a surface, functions that are good at surface fitting could provide a potential solution for facial pattern mathematics learning. Focusing on data approximation, multi-Gaussian function is considered highly efficient for representing a relatively complex waveform or surface using a very small set of parameters and has been successfully applied to fit many waveforms and surfaces. It can therefore be speculated that by combining the multi-Gaussian function with prior knowledge of face characteristics, it is possible to construct a mathematical model of a face image. No such research has been attempted at this stage and the feasibility of this method requires further verification.

Consequently, unlike studies driven by data or task, the goal of this work is to provide a mathematical model for face image representation. The proposed approach will enable the transformation of a face image by a series of parameters, and will likely provide valuable insights into the facial systems being modeled. It is also expected to provide more convenient and effective means for the applications of image and animation processing technologies such as face synthesis, bionic expression, and cartoon character design.

The main contributions of this work can be summarized as follows:
- A mathematical model (GmFace) for face image representation is proposed in the form of a multi-Gaussian function.
- A neural network (GmNet) is designed for parameter solving of multi-Gaussian function.
- A modeling method for a face image with GmFace is presented, including common face modeling and personal face modeling.
- While the GmFace is only an approximate (not a perfect or the simplest) model for face image representation, it is the first mathematical face model in an explicit function form with good image transformation and representation abilities.

The remainder of this paper is organized as follows. Related works are discussed in Section 2. The proposed method is formulated in Section 3. Comprehensive experiments demonstrating the effectiveness of the proposed method are presented in Section 4, and finally, Section 5 provides a conclusion with a summary and outlook for future work.

## II. RELATED WORK

The mathematical learning of face morphological structure belongs to the research category of complex biological system modeling. In principle, as a simplified representation of reality, such models should both facilitate understanding of biological systems and also predict the consequences of interventions on biological systems [24], [25]. T. Melham et al. (2013) used the essence of this 'representation' relationship to correlate model states and trajectories with experimental observations [26]. According to this research, facial mathematical modeling is a process of simplifying 'representation' and must reflect the correlation between model parameters and face image to a certain extent.

In practical applications, Gaussian function is one of the most important elementary functions that is widely used in mathematical analysis, image processing, and engineering modeling. It is well known that the Gaussian function is integrable and differentiable, and its characteristic bell-shaped graph appears everywhere from normal distribution in statistics to position wave packets of a particle in quantum mechanics [27]. Based on multi-Gaussian function, waveform fitting has shown remarkable success in a wide range of applications including physiological signals representation [28], [29], [30], [31], [32], pulsar signal simulation [33], [34], [35], and human pose detection [36]. L. Wang et al. proposed a multi-Gaussian model which closely approximates a single-period digital volume pulse signal by decomposing it into four or five Gaussian waves, in which the quantity of Gaussian function is determined by the morphology of the pulse waveform [28]. H. Zhang et al. employed multi-Gaussian functions to express an X-ray pulsar profile. It has been confirmed that in addition to the physical meaning, the multi-Gaussian fitting method can accurately describe the structure and minutiae of the pulsar profile [33]. Moreover, multi-Gaussian-based methods are also applied for fitting image silhouettes. R. Y. Xu et al. introduced multi-Gaussian function into ellipse fitting and successfully completed the test on an image silhouette of a human upper body [36]. Those studies focused on representing a one-dimensional (1D) waveform curve by a combination of Gaussian waves. For the two-dimensional (2D) surface, multi-Gaussian has been used for surface fitting scattered data as early as 1993. A. Goshtasby et al. designed a sum of Gaussians to approximate 100 scattered points. As this research implies, multi-Gaussian function is more effective at dealing with a large data set from a highly varying surface [37]. However, limited by computing resources at the time, providing a larger parameter space was a serious problem, making it very difficult to represent complex surfaces like the human face. With the development of computer science, the high-performance of computing resources and the efficiency of neural network algorithms provide optimal conditions for solving the parameters in a very large space. Therefore, multi-Gaussian function is considered as an ideal option to provide an approximate mathematical expression of a face image.

Based on the above analysis, multi-Gaussian function is employed to provide an explicit mathematical representation of a face image in this work. Furthermore, a GmNet architecture is designed to solve the estimation of parameters, thereby enabling the face image to be represented mathematically. To the best of the authors' knowledge, this is the first work to construct a mathematical model of a human frontal face image.

## III. METHODS

### A. Face Image Model: GmFace

A face image is a projection of a human face onto a 2D plane. In the usual analysis of face images, pixel intensities are the most common and easiest exploited factors [38], [39], [40], [41], [42]. However, the spatial coordinates and relations between pixel intensities and spatial coordinates are often ignored. To remedy this, the face image is represented as a 2D surface, and two dimensional spatial coordinates are taken as independent variables and coordinate-based pixel intensities as dependent variables.

It is well known that Gaussian function is a complete set on $L^2(R^n)$ [43], which is stated mathematically as follows:

Let $f(\cdot): R^n \to R$ be any nonnegative real integrable function on $R^n$. An approximation of $f(\cdot)$ then exists by linear combinations of multi-Gaussian function as the form:

$$\hat{f}(\mathbf{x}) = \sum_{i=1}^{m} w_i G_i(\mathbf{x}, \theta_i) \quad (1)$$

where $\mathbf{x}$ is the input, $G_i$ is a Gaussian function, $\theta_i$ denotes the parameters of Gaussian function $G_i$, $w_i$ is the weight coefficient, and $m$ is the quantity of Gaussian components.

In a 2D space, the bell-shaped graph of Gaussian function appears as a bell surface, and its spatial position, size and direction can be controlled by a set of parameters [44]. Most importantly, Gaussian function is a complete set on $L^2(R^n)$, meaning that the finite multi-Gaussian function can approximate any non-negative integrable functions on a real number with arbitrary accuracy. Based on these considerations, multi-Gaussian function is an ideal option to provide the mathematical expression of GmFace for face image representation approximatively.

Thus, for each pixel, the GmFace model can be constructed as:

$$GmFace(x_1, x_2) = \sum_{i=1}^{m} w_i G_i(x_1, x_2 \mid \boldsymbol{\mu}_i, \mathbf{A}_i) \quad (2)$$

Here, a face image is expressed as a linear combination of a series of Gaussian components, where $w_i$ is the weight coefficient, $m$ represents the quantity of Gaussian components, and $G_i$ is a Gaussian function as:

$$G(\mathbf{x} \mid \boldsymbol{\mu}, \mathbf{A}) = \exp\{-(\mathbf{x}-\boldsymbol{\mu})^T \mathbf{A}(\mathbf{x}-\boldsymbol{\mu})\} \quad (3)$$

where $\mathbf{x} = \begin{bmatrix} x_1 \\ x_2 \end{bmatrix}$ is the input formed with 2D spatial coordinates, $\boldsymbol{\mu} = \begin{bmatrix} \mu_1 \\ \mu_2 \end{bmatrix}$ denotes the Gaussian center, and $\mathbf{A} = \begin{bmatrix} a_{11} & a_{12} \\ a_{21} & a_{22} \end{bmatrix}$ is a positive-definite symmetric matrix named the precision matrix, which is the inverse matrix of covariance matrix.

### B. Multi-Gaussian Network: GmNet

As the pixel points of a face image are huge in size and the face has a complex physiological structure, estimating the parameters of GmFace model is a difficult task. One of the most challenging issues in parameter estimations for GmFace is the large amount of calculation required. To determine the complex parameters, a 3-layer neural network named GmNet is constructed, in which each single Gaussian function is used as a neuron. The network structure is illustrated in Fig. 1.

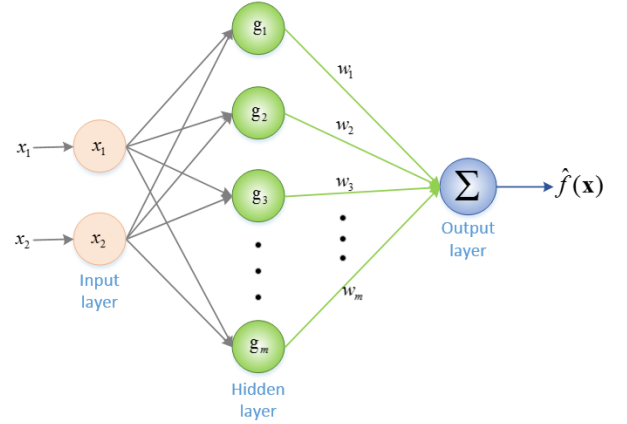

Fig. 1. The network structure of GmNet.

The input layer of the framework of GmNet is the 2D surface1 data of face images and the hidden layer is a group of Gaussian modules truncated to a region bounded by the image size. Eventually, in the output layer, the response maps are combined by linear weighting $w$. The weight coefficient $w_i$ summarizes the contribution of the ith Gaussian kernel to the output.

In the process of solving the GmFace parameters through GmNet, it is necessary to guarantee the positive definiteness of $\mathbf{A}$ in Eq. (3) and certain constraints must therefore be considered. According to the properties of positive semidefinite symmetric matrix, the positive definite of $\mathbf{A}$ is equivalent to that it has a unique Cholesky decomposition [45]:

$$\mathbf{A} = \mathbf{L}\mathbf{L}^T \quad (4)$$

where $\mathbf{L}$ is a lower triangular matrix with real and strictly positive diagonal elements. According to the above decomposition, Eq. (3) can be derived as:

$$G(\mathbf{x} \mid \boldsymbol{\mu}, \mathbf{L}) = \exp\{-(\mathbf{x}-\boldsymbol{\mu})^T \mathbf{L}\mathbf{L}^T (\mathbf{x}-\boldsymbol{\mu})\} \quad (5)$$

Accordingly, GmFace as Eq. (2) is rewritten as:

$$GmFace(x_1, x_2) = \sum_{i=1}^{m} w_i G_i(x_1, x_2 \mid \boldsymbol{\mu}_i, \mathbf{L}_i) \quad (6)$$

where $\mathbf{x} = \begin{bmatrix} x_1 \\ x_2 \end{bmatrix} = \begin{bmatrix} \dfrac{r}{H} \\ \dfrac{c}{W} \end{bmatrix}$ is the row and column coordinates of image after normalization, $r$ and $c$ are the row index and column index, respectively, and $W$ and $H$ are the width and height of a face image, respectively.

Thus, in GmNet, $m$ groups of $\boldsymbol{\mu}$ and $\mathbf{L}$ are set to generate truncated Gaussian components, and each component size is the same as the image size. The $w_i$, $\boldsymbol{\mu}_i$ and $\mathbf{L}_i$ are the parameters that must be solved.

For the purpose of optimization solution, the GmNet is trained with error back-propagation algorithm [46]. During training, the positive definiteness of precision matrix $\mathbf{A}$ is guaranteed by strictly constraining the positive diagonal elements of $\mathbf{L}$ and its lower triangle matrix form.

---

[1] In order to provide a more readable visualization effect, the 2D surface of face images expressed in the figures of this paper consists of the value of "1-GmFace".

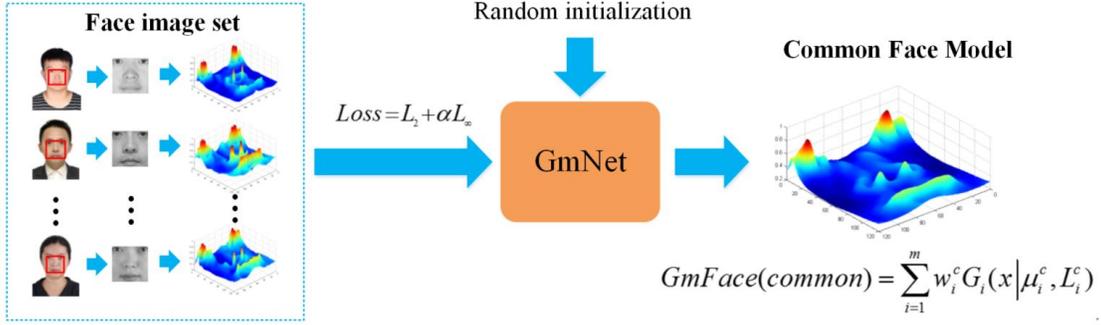

Fig. 2. The process of common face modeling. With a group training face images are fed to GmNet as targets, GmNet can estimate the general representation of face pattern and obtain parameters that can express face common features.

By applying the chain rule, the partial derivatives of *GmFace* can be obtained with respect to all parameters as follows:

$$\frac{\partial GmFace}{\partial w_i} = G(\mathbf{x}|\boldsymbol{\mu}_i, \mathbf{L}_i) \quad (7)$$

$$\frac{\partial GmFace}{\partial \boldsymbol{\mu}_i} = \frac{\partial GmFace}{\partial G} * \frac{\partial G}{\partial \boldsymbol{\mu}_i} = w_i G(\mathbf{x}|\boldsymbol{\mu}_i, \mathbf{L}_i) \mathbf{L}_i \mathbf{L}_i^T (\mathbf{x} - \boldsymbol{\mu}) \quad (8)$$

$$\frac{\partial GmFace}{\partial \mathbf{L}_i} = \frac{\partial GmFace}{\partial G} * \frac{\partial G}{\partial \mathbf{L}_i}$$
$$= -w_i G(\mathbf{x}|\boldsymbol{\mu}_i, \mathbf{L}_i)(\mathbf{x} - \boldsymbol{\mu}_i)(\mathbf{x} - \boldsymbol{\mu}_i)^T \mathbf{L}_i \quad (9)$$

Two measurements are designed to indicate the global and local differences between the built model and the objective function as follows:

$$L_2 = \sum (GmFace(\mathbf{x}|\boldsymbol{\mu}_i, \mathbf{L}_i) - f(\mathbf{x}))^2 \quad (10)$$

$$L_\infty = \max(|GmFace(\mathbf{x}|\boldsymbol{\mu}_i, \mathbf{L}_i) - f(\mathbf{x})|) \quad (11)$$

A scaling factor $\alpha$ is then set for balancing the two measurements and the loss function used for optimizing parameters of GmNet is constructed as:

$$Loss = L_2 + \alpha L_\infty \quad (12)$$

Adaptive moment estimation (Adam) is a gradient descent optimization algorithm [47], [48] which computes adaptive learning rates for each parameter. Equation (7) to (9) indicate that parameters of GmFace model have very different derivative forms. In GmNet, Adam algorithm is used to update each parameter to ensure the magnitudes of parameter updates are invariant to rescaling of the gradient.

### C. Common Face Modeling

The process of common face modeling is constructed as illustrated in Fig. 2. To estimate the general representation of face pattern and obtain parameters of common face model, a group of training face images are fed to GmNet as targets.

Specifically, the loss function in Eq. (12) can be rewritten as Eq. (13), and the difference between model output and training face images is measured as:

$$Loss = L_2 + \alpha L_\infty$$
$$= \sum_j^N \sum_{x_1=\frac{1}{H}}^1 \sum_{x_1=\frac{1}{W}}^1 (GmFace(x_1, x_2|\boldsymbol{\mu}_i, \mathbf{L}_i) - f_j(x_1, x_2))^2 \quad (13)$$
$$+ \alpha \max_{j \in [1,N], x_1 \in [\frac{1}{H},1], x_2 \in [\frac{1}{W},1]} |GmFace(x_1, x_2|\boldsymbol{\mu}_i, \mathbf{L}_i) - f_j(x_1, x_2)|$$

where $L_2$ denotes mean square error (MSE) and $L_\infty$ is the proposed loss function peak absolute error (PAE). The $W$ and $H$ are the width and height of a face image, respectively, and $N$ is the number of training face images.

In this way, the global error is taken into account by MSE, as well as the local error, which can be measured by PAE. Finally, the GmFace parameters for common feature representation are obtained by minimizing the loss function.

### D. Personal Face Modeling

The implementation details of personal face modeling are shown in Fig. 3. Face image representation is the process of transforming face information from the image pixel space to the parameter space of GmFace. Learning through a large number of face images, the proposed GmFace model gives an approximate mathematical expression of the projection space of the human face, which reflects the common features. Each personal face image can be regarded as a point in the space which needs to further express the individual characteristics based on common face features.

For the mathematical expression of a personal face image, the loss function is used to measure the difference between model output and the face image to be represented, which is denoted as:

$$Loss = L_2 + \alpha L_\infty$$
$$= \sum_{x_1=\frac{1}{H}}^1 \sum_{x_2=\frac{1}{W}}^1 (GmFace(x_1, x_2|\boldsymbol{\mu}_i, \mathbf{L}_i) - f(x_1, x_2))^2 \quad (14)$$
$$+ \alpha \max_{x_1 \in [\frac{1}{H},1], x_2 \in [\frac{1}{W},1]} |GmFace(x_1, x_2|\boldsymbol{\mu}_i, \mathbf{L}_i) - f(x_1, x_2)|$$

The 2D surface of the specific face image is considered the learning target and the GmNet is established to optimize the parameters. The solved parameters of common face model are then used for the initialization of personal face modeling.

Finally, based on the pattern learning of common features, the mathematical expression of the personal face image can be written explicitly after solving the personalized parameters.

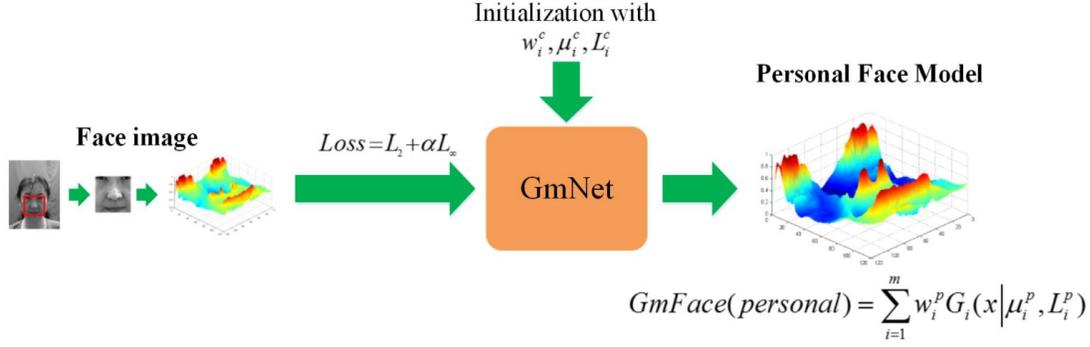

Fig. 3. The process of personal face modeling. The 2D surface of the specific face image is considered as the learning target and initialized with common face model, GmNet can solve the personalized parameters, and the mathematical expression of the human face image can be written explicitly through the solved parameters

### E. Face Image Transformation through GmFace

By applying GmFace model, the face image transformation becomes much simpler. The description of facial features is presented by the model parameters. In this way, the complex image processing is replaced by a simple mathematical calculation. The following is the specific derivation of the transformation operation using GmFace model.

#### 1) Image Translation

The image translation by GmFace is implemented by adjusting parameter $\mu_i$ as:

$$GmFace(\mathbf{x} - \bar{\mathbf{x}})$$
$$= \sum_{i=1}^{m} w_i \exp\{-(\mathbf{x} - \bar{\mathbf{x}} - \mu_i)^T \mathbf{A}_i (\mathbf{x} - \bar{\mathbf{x}} - \mu_i)\}$$
$$= \sum_{i=1}^{m} w_i \exp\{-(\mathbf{x} - (\mu_i + \bar{\mathbf{x}}))^T \mathbf{A}_i (\mathbf{x} - (\mu_i + \bar{\mathbf{x}}))\} \quad (15)$$
$$= \sum_{i=1}^{m} w_i \exp\{-(\mathbf{x} - \bar{\mu}_i)^T \mathbf{A}_i (\mathbf{x} - \bar{\mu}_i)\}$$

where $\bar{\mathbf{x}} = \begin{bmatrix} x_1 \\ x_2 \end{bmatrix}$ is the translational vector and $\bar{\mu}_i = \mu_i + \bar{\mathbf{x}}$.

#### 2) Image Scaling

Image scaling by GmFace is implemented by adjusting the parameters $\mu_i$ and $\mathbf{A}_i$ as:

$$GmFace(k\mathbf{x})$$
$$= \sum_{i=1}^{m} w_i \exp\{-(k\mathbf{x} - \mu_i)^T \mathbf{A}_i (k\mathbf{x} - \mu_i)\}$$
$$= \sum_{i=1}^{m} w_i \exp\{-(\mathbf{x} - \frac{1}{k}\mu_i)^T (k^2 \mathbf{A}_i)(\mathbf{x} - \frac{1}{k}\mu_i)\} \quad (16)$$
$$= \sum_{i=1}^{m} w_i \exp\{-(\mathbf{x} - \bar{\mu}_i)^T \bar{\mathbf{A}}_i (\mathbf{x} - \bar{\mu}_i)\}$$

where $k$ is the scaling factor and $\bar{\mu}_i = \frac{1}{k}\mu_i$, $\bar{\mathbf{A}}_i = k^2 \mathbf{A}_i$.

#### 3) Image Rotation

The image rotation around any point by GmFace is implemented by adjusting the parameter $\mu_i$ and $\mathbf{A}_i$ as:

$$GmFace((\mathbf{F}_r \cdot (\mathbf{x} - \bar{\mathbf{x}})) + \bar{\mathbf{x}})$$
$$= \sum_{i=1}^{m} w_i \exp\{-(\mathbf{F}_r \cdot (\mathbf{x} - \bar{\mathbf{x}}) - \mu_i + \bar{\mathbf{x}})^T$$
$$\cdot \mathbf{A}_i \cdot (\mathbf{F}_r \cdot (\mathbf{x} - \bar{\mathbf{x}}) - \mu_i + \bar{\mathbf{x}})\}$$
$$= \sum_{i=1}^{m} w_i \exp\{-(\mathbf{x} - \mathbf{F}_r^{-1} \cdot (\mu_i + \mathbf{F}_r \cdot \bar{\mathbf{x}} - \bar{\mathbf{x}}))^T \quad (17)$$
$$\cdot \mathbf{F}_r^T \mathbf{A}_i \mathbf{F}_r \cdot (\mathbf{x} - \mathbf{F}_r^{-1} \cdot (\mu_i + \mathbf{F}_r \cdot \bar{\mathbf{x}} - \bar{\mathbf{x}}))\}$$
$$= \sum_{i=1}^{m} w_i \exp\{-(\mathbf{x} - \bar{\mu}_i)^T \bar{\mathbf{A}}_i \cdot (\mathbf{x} - \bar{\mu}_i)\}$$

where $\mathbf{F}_r = \begin{bmatrix} \cos\theta & \sin\theta \\ -\sin\theta & \cos\theta \end{bmatrix}$ denotes the rotation matrix, $\theta$ is the rotation angle, $\bar{\mathbf{x}} = \begin{bmatrix} x_1 \\ x_2 \end{bmatrix}$ is the center of rotation, and $\bar{\mu}_i = \mathbf{F}_r^{-1} \cdot (\mu_i + \mathbf{F}_r \cdot \bar{\mathbf{x}} - \bar{\mathbf{x}})$, $\bar{\mathbf{A}}_i = \mathbf{F}_r^T \mathbf{A}_i \mathbf{F}_r$.

## IV. EXPERIMENTS

### A. Experimental Setup

#### 1) Data Sets

The study sample for common face modeling was comprised of 1870 individual Chinese face images obtained from a frontal perspective. The gender and age distribution are summarized in Table 1.

In the experiment of personal face modeling, 1040 frontal normal images from Chinese face database CAS-PEAL-R1 [49] were used to validate the effectiveness of GmFace. In addition, images from CAS-PEAL-R1 were not involved in the computing of the common face model.

To reduce the influence of background, the face image samples were cropped and adjusted to pixels in the pre-processing. The image size was 120×120 pixels and a selection of data examples are provided in Fig. 4.

The index of the assessment is the reconstruction MSE and parameter size of each observed method.

For GmFace model, the parameter size is:

$$Parameter\_Size_{GmFace} = 6 \times m \quad (18)$$

Here, $m$ represents the quantity of Gaussian components in multi-Gaussian function, 6 is the parameter size of each 2D Gaussian component, including the $\mu$ vector of two variables, symmetric matrix $\mathbf{A}$ of three variables, and weight coefficient

TABLE 1
CHARACTERISTICS OF THE STUDY SAMPLE

| Factors | Group | Quantity | Percent |
|---|---|---|---|
| Gender | Female | 724 | 38.72% |
|  | Male | 1146 | 61.28% |
| Age | 18-29 | 399 | 21.34% |
|  | 30-39 | 394 | 21.07% |
|  | 40-49 | 443 | 23.69% |
|  | 50-59 | 360 | 19.25% |
|  | ≥60 | 274 | 14.65% |

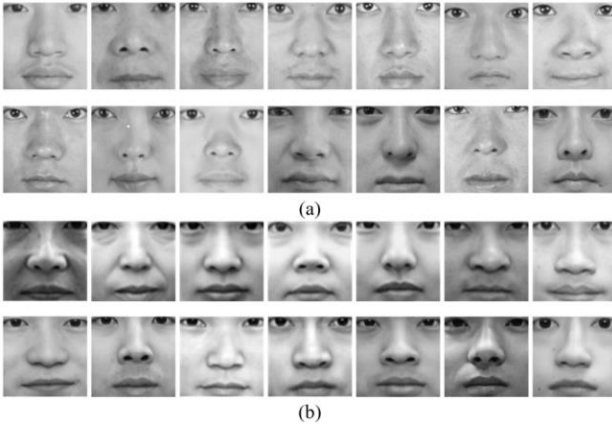

Fig. 4. Face image examples for GmFace modeling: (a) Image examples with background removal and rescaling for solving common face model. (b) Image examples from the CAS-PEAL-R1 database for the verification of personal face model.

$w$.

*2) Parameter Setting*

The experiments were carried out on the open-source deep learning toolkit PyTorch [50]. During common face modeling, the batch size was set to 256, and hyperparameters in Adam optimization algorithm were set to default values (refer to [47], [48]). The parameters of GmFace were solved through GmNet.

*B. Experimental Results*

*1) Visualized Results of 2D Surface for a Face Image*

In GmFace modeling, a face image is expressed as a 2D surface, which takes 2D spatial coordinates as independent variables and pixel intensities as dependent variables. The observation of the proposed 2D surface is shown in Fig. 5.

As illustrated, faces have a unique pattern that is a continuously changing surface. In particularly, there are more obvious peak-valley distributions at the eyes, nose, and mouth, while the cheek is more gentle and smooth. According to the visualization results, the hypothesis of face modeling using multi-Gaussian function is verified. As the curve shapes show, it appears that each curve shape is a segment truncated of multi-Gaussian function by a sliding window with the image size.

*2) Modeling Results of Common Face*

The visualization of proposed GmNet output for common face modeling are demonstrated in Fig. 6. It can be intuitively observed that with an increase in the number of epochs, GmNet obtains a superior performance in common face representation

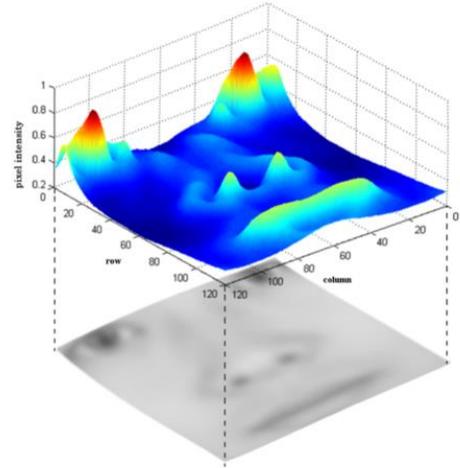

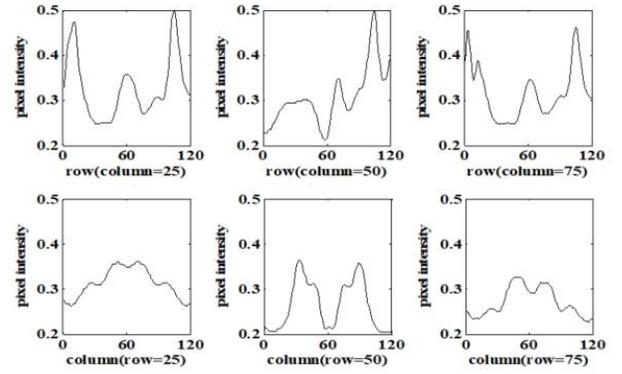

Fig. 5. The 2D surface of a face image. (a) 2D surface of average face image in gray level with 120×120 pixels image size. (b) Six vertical cross sections of the 2D surface of average face image, among which half are perpendicular to the x-axis with a label "row" and half are perpendicular to the y-axis with a label "column".

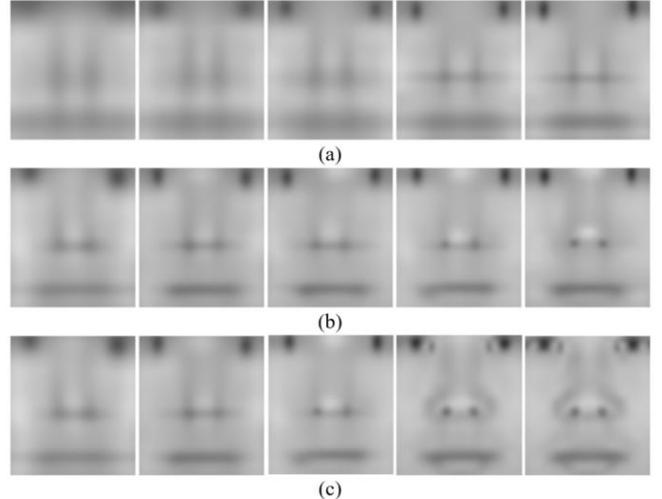

Fig. 6. Face pattern modeling with common features based on GmFaceNet. (a) The outputs changing with different number of epochs under 40 Gaussian components. (b) The outputs changing with different number of epochs under 60 Gaussian components. (c) The outputs changing with different number of epochs under 80 Gaussian components. From left to right in all sub-figures ((a)&(b)&(c)), the corresponding numbers of epochs are 100, 200, 500, 1000, 2000 respectively.

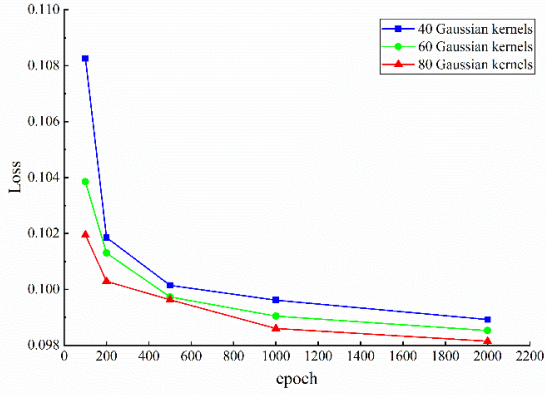

Fig. 7. Loss curve of GmFaceNet with different Gaussian kernel numbers.

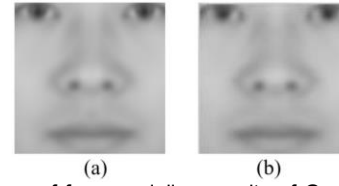

Fig. 8. Comparison of face modeling results of GmFace with average face. (a) General face image constructed by GmFace model. (b) Average face image computed from training data.

TABLE 2
CORRESPONDING LOSSES OF THREE GMNET ARCHITECTURES WITH DIFFERENT QUANTITY OF GAUSSIAN COMPONENTS

| | Loss | | | | |
|---|---|---|---|---|---|
| $m$ | epoch= 100 | epoch= 200 | epoch= 500 | epoch= 1000 | epoch= 2000 |
| 40 | 0.10826 | 0.10185 | 0.10014 | 0.09962 | 0.09892 |
| 60 | 0.10385 | 0.10130 | 0.09973 | 0.09905 | 0.09853 |
| 80 | 0.10195 | 0.10029 | 0.09963 | 0.09860 | 0.09815 |

*m is the quantity of Gaussian components.*

and the optimization method of error back-propagation is suitable for the parameter solution of GmFace model through GmNet. For the parameter $m$, which is the quantity of Gaussian components in GmFace, the common face model provides a better representation when is 80 compared to when it is 40 and 60, and the output common face model is already recognizable as a face pattern with the naked eye.

The corresponding losses of three GmNet architectures with different quantity of Gaussian components (40, 60, and 80) are illustrated in Fig. 7 and in Table 2.

The GmFace model with 80 Gaussian components can be observed to obtain the minimal loss, providing further evidence that the GmFace model with 80 Gaussian components provides a superior representation. A total of 480 GmFace model parameters are solved through GmNet and can be used for the initialization of personal face modeling, as listed in Table 3.

It should be noted that the parametric solution for GmFace model is not unique. Under different initialization conditions, the parameters solved by the GmNet may also be different and this does not affect the utilization of the models. No matter which solution group is used, it is acceptable as long as it meets the task expectations and goals.

The face image constructed by GmFace model and the average face image computed from training data set are illustrated in Fig. 8. The common face model fitted by GmFace can be observed to provide the same visual effect as the average face. To further evaluate the GmFace fitting results, the MSE between these two face images was calculated (gray values are normalized to [0, 1]) and the obtained result is MSE = $2.86e^{-5}$. Therefore, according to the results of the experiments, the

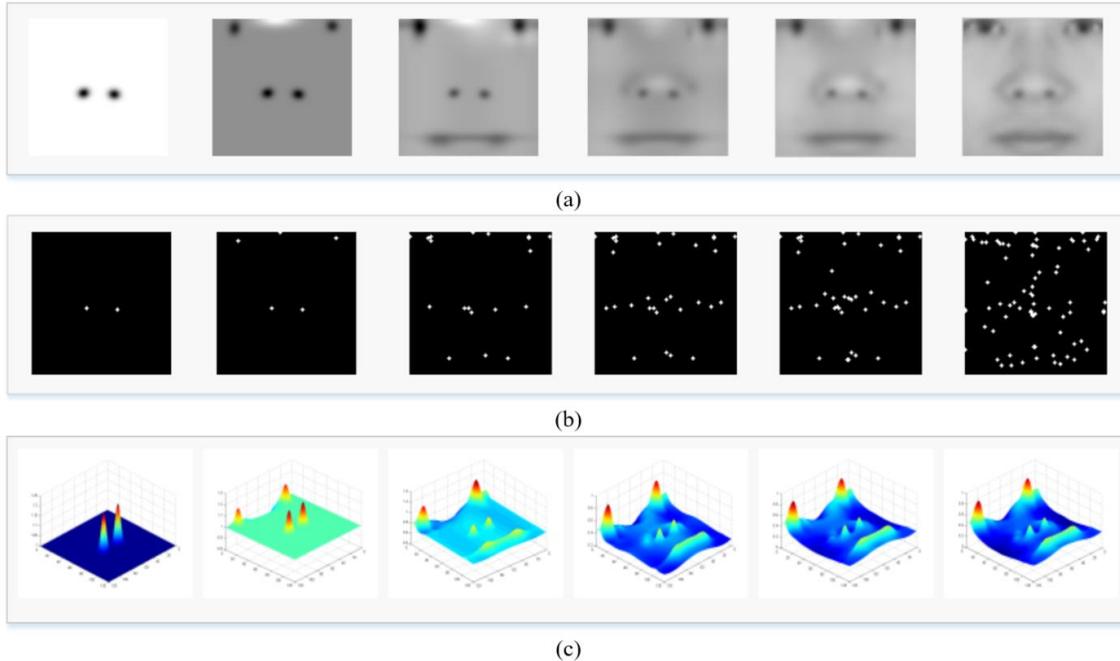

Fig. 9. Face model analysis based on GmFace. (a) Outputs of the model with k Gaussian components with weight from highest to lowest. (b) Positions of the center of the k Gaussian components. (c) The 2D surface of outputs with k Gaussian components according to row and column coordinates. In all sub-figures ((a),(b), and (c)), $k$ is equal to 2, 5, 20, 40, 60, and 80 from left to right, respectively.

TABLE 3
A GROUP OF SOLUTIONS OF GMFACE MODEL PARAMETERS WITH 80 GAUSSIAN COMPONENTS FOR COMMON FACE MODEL

| Parameter | Value | | | | | | | |
|---|---|---|---|---|---|---|---|---|
| $i$ | 1 | 2 | 3 | 4 | 5 | 6 | 7 | 8 |
| $w_i$ | 0.1800 | 0.1154 | -0.0769 | -0.0634 | 0.0780 | 0.0927 | 0.0774 | 0.1803 |
| $\mu_i$ | $\begin{bmatrix}0.0592\\0.0641\end{bmatrix}$ | $\begin{bmatrix}0\\0.2762\end{bmatrix}$ | $\begin{bmatrix}0.3425\\0.0704\end{bmatrix}$ | $\begin{bmatrix}0.8814\\0.2777\end{bmatrix}$ | $\begin{bmatrix}0.1525\\0.1305\end{bmatrix}$ | $\begin{bmatrix}0.5548\\0.5114\end{bmatrix}$ | $\begin{bmatrix}0.4991\\0.4422\end{bmatrix}$ | $\begin{bmatrix}0.0845\\0.4927\end{bmatrix}$ |
| $A_i$ | $\begin{bmatrix}513.51 & -75.36\\-75.36 & 1226.44\end{bmatrix}$ | $\begin{bmatrix}51.95 & -62.85\\-62.85 & 219.40\end{bmatrix}$ | $\begin{bmatrix}67.10 & -3.78\\-3.78 & 51.51\end{bmatrix}$ | $\begin{bmatrix}487.19 & -188.87\\-188.87 & 225.17\end{bmatrix}$ | $\begin{bmatrix}150.29 & 18.67\\18.67 & 78.31\end{bmatrix}$ | $\begin{bmatrix}0.0161 & 0.0003\\0.0003 & 0.0029\end{bmatrix}$ | $\begin{bmatrix}0.0163 & -0.0007\\-0.0007 & 0.0736\end{bmatrix}$ | $\begin{bmatrix}173.79 & 21.13\\21.13 & 46.80\end{bmatrix}$ |
| $i$ | 9 | 10 | 11 | 12 | 13 | 14 | 15 | 16 |
| $w_i$ | 0.0204 | -0.1428 | -0.2473 | 0.0554 | 0.0484 | -0.1474 | 0.1523 | -0.2013 |
| $\mu_i$ | $\begin{bmatrix}0.4425\\0.5051\end{bmatrix}$ | $\begin{bmatrix}0.0698\\0.1508\end{bmatrix}$ | $\begin{bmatrix}0.5434\\0.3917\end{bmatrix}$ | $\begin{bmatrix}0.8164\\0.5454\end{bmatrix}$ | $\begin{bmatrix}0.9445\\0.2449\end{bmatrix}$ | $\begin{bmatrix}0.5725\\0.5049\end{bmatrix}$ | $\begin{bmatrix}0.0070\\0.7453\end{bmatrix}$ | $\begin{bmatrix}0.0594\\0.1971\end{bmatrix}$ |
| $A_i$ | $\begin{bmatrix}0.0190 & -0.0154\\-0.0154 & 40.6306\end{bmatrix}$ | $\begin{bmatrix}391.88 & 66.77\\66.77 & 1141.50\end{bmatrix}$ | $\begin{bmatrix}629.45 & 56.35\\56.35 & 464.89\end{bmatrix}$ | $\begin{bmatrix}10.0140 & -0.0093\\-0.0093 & 0.3229\end{bmatrix}$ | $\begin{bmatrix}56.01 & -8.93\\-8.93 & 27.52\end{bmatrix}$ | $\begin{bmatrix}327.14 & 15.71\\15.71 & 104.30\end{bmatrix}$ | $\begin{bmatrix}513.30 & 13.57\\13.57 & 2.54\end{bmatrix}$ | $\begin{bmatrix}643.43 & 168.79\\168.79 & 138.31\end{bmatrix}$ |
| $i$ | 17 | 18 | 19 | 20 | 21 | 22 | 23 | 24 |
| $w_i$ | -0.1346 | 0.2281 | 0.0798 | -0.0747 | -0.1219 | 0.0932 | 0.1110 | -0.0848 |
| $\mu_i$ | $\begin{bmatrix}0.8567\\0.3827\end{bmatrix}$ | $\begin{bmatrix}0.0667\\0.2260\end{bmatrix}$ | $\begin{bmatrix}0.4778\\0.1575\end{bmatrix}$ | $\begin{bmatrix}0.3123\\0.5725\end{bmatrix}$ | $\begin{bmatrix}0.0363\\0.0239\end{bmatrix}$ | $\begin{bmatrix}0.4954\\0.4231\end{bmatrix}$ | $\begin{bmatrix}0.2812\\0.4985\end{bmatrix}$ | $\begin{bmatrix}0.5729\\0.3017\end{bmatrix}$ |
| $A_i$ | $\begin{bmatrix}336.63 & 62.86\\62.86 & 85.12\end{bmatrix}$ | $\begin{bmatrix}1130.73 & 383.50\\383.50 & 1518.67\end{bmatrix}$ | $\begin{bmatrix}0.0554 & 0.0176\\0.0176 & 17.5398\end{bmatrix}$ | $\begin{bmatrix}144.05 & 5.40\\5.40 & 407.53\end{bmatrix}$ | $\begin{bmatrix}1960.47 & 1076.84\\1076.84 & 897.95\end{bmatrix}$ | $\begin{bmatrix}0.0919 & 0.0187\\0.0187 & 38.3043\end{bmatrix}$ | $\begin{bmatrix}62.2562 & 0.0194\\0.0194 & 0.0063\end{bmatrix}$ | $\begin{bmatrix}446.69 & -257.68\\-257.68 & 356.56\end{bmatrix}$ |
| $i$ | 25 | 26 | 27 | 28 | 29 | 30 | 31 | 32 |
| $w_i$ | 0.0103 | -0.1224 | -0.0415 | 0.0842 | 0.0489 | -0.0971 | -0.1081 | 0.0388 |
| $\mu_i$ | $\begin{bmatrix}0.5377\\0.1809\end{bmatrix}$ | $\begin{bmatrix}0.4581\\0.3049\end{bmatrix}$ | $\begin{bmatrix}0.9434\\0.3494\end{bmatrix}$ | $\begin{bmatrix}0.8371\\0.2879\end{bmatrix}$ | $\begin{bmatrix}1.0000\\0.2187\end{bmatrix}$ | $\begin{bmatrix}0.1804\\0.6297\end{bmatrix}$ | $\begin{bmatrix}0.8630\\0.5244\end{bmatrix}$ | $\begin{bmatrix}0.6796\\0.2038\end{bmatrix}$ |
| $A_i$ | $\begin{bmatrix}121.69 & 75.27\\75.27 & 562.65\end{bmatrix}$ | $\begin{bmatrix}270.31 & 137.76\\137.76 & 178.04\end{bmatrix}$ | $\begin{bmatrix}838.67 & -240.47\\-240.47 & 243.79\end{bmatrix}$ | $\begin{bmatrix}201.81 & 53.51\\53.51 & 124.86\end{bmatrix}$ | $\begin{bmatrix}507.21 & -15.32\\-15.32 & 149.22\end{bmatrix}$ | $\begin{bmatrix}4.49 & -6.26\\-6.26 & 303.54\end{bmatrix}$ | $\begin{bmatrix}792.67 & 45.41\\45.41 & 35.24\end{bmatrix}$ | $\begin{bmatrix}162.91 & 51.93\\51.93 & 141.52\end{bmatrix}$ |
| $i$ | 33 | 34 | 35 | 36 | 37 | 38 | 39 | 40 |
| $w_i$ | 0.0674 | 0.0668 | -0.0407 | -0.1333 | 0.1335 | -0.0419 | 0.1017 | -0.2504 |
| $\mu_i$ | $\begin{bmatrix}0.1989\\0.3294\end{bmatrix}$ | $\begin{bmatrix}0.6528\\0.0924\end{bmatrix}$ | $\begin{bmatrix}0.8821\\0.7841\end{bmatrix}$ | $\begin{bmatrix}0.0877\\0.2824\end{bmatrix}$ | $\begin{bmatrix}0.6295\\0.0059\end{bmatrix}$ | $\begin{bmatrix}0.5607\\0.0375\end{bmatrix}$ | $\begin{bmatrix}0.5427\\0.5525\end{bmatrix}$ | $\begin{bmatrix}0.0227\\0.1202\end{bmatrix}$ |
| $A_i$ | $\begin{bmatrix}228.62 & 30.49\\30.49 & 43.65\end{bmatrix}$ | $\begin{bmatrix}43.63 & 1.87\\1.87 & 68.28\end{bmatrix}$ | $\begin{bmatrix}496.21 & -209.76\\-209.76 & 218.91\end{bmatrix}$ | $\begin{bmatrix}801.00 & 103.64\\103.64 & 1101.65\end{bmatrix}$ | $\begin{bmatrix}0.0167 & -0.0193\\-0.0193 & 52.6096\end{bmatrix}$ | $\begin{bmatrix}23.45 & -17.76\\-17.76 & 151.13\end{bmatrix}$ | $\begin{bmatrix}0.0013 & 0.0004\\0.0004 & 0.6579\end{bmatrix}$ | $\begin{bmatrix}387.74 & 118.07\\118.07 & 94.99\end{bmatrix}$ |
| $i$ | 41 | 42 | 43 | 44 | 45 | 46 | 47 | 48 |
| $w_i$ | 0.0843 | 0.0514 | -0.0524 | -0.1318 | 0.1256 | 0.0925 | 0.0705 | 0.0735 |
| $\mu_i$ | $\begin{bmatrix}0.0711\\0.9500\end{bmatrix}$ | $\begin{bmatrix}0\\0.4258\end{bmatrix}$ | $\begin{bmatrix}0.3712\\0.9342\end{bmatrix}$ | $\begin{bmatrix}0.9108\\0.6954\end{bmatrix}$ | $\begin{bmatrix}0.0515\\0.9387\end{bmatrix}$ | $\begin{bmatrix}0.5569\\0.4594\end{bmatrix}$ | $\begin{bmatrix}0.4300\\0.4411\end{bmatrix}$ | $\begin{bmatrix}0.1420\\0.8216\end{bmatrix}$ |
| $A_i$ | $\begin{bmatrix}693.10 & 635.93\\635.93 & 990.25\end{bmatrix}$ | $\begin{bmatrix}14838.52 & 5.42\\5.42 & 0.0125\end{bmatrix}$ | $\begin{bmatrix}74.01 & -31.26\\-31.26 & 181.55\end{bmatrix}$ | $\begin{bmatrix}336.43 & 116.36\\116.36 & 173.63\end{bmatrix}$ | $\begin{bmatrix}870.24 & 432.65\\432.65 & 2753.05\end{bmatrix}$ | $\begin{bmatrix}0.0020 & -0.0008\\-0.0008 & 0.2259\end{bmatrix}$ | $\begin{bmatrix}0.2846 & 0.0148\\0.0148 & 0.0567\end{bmatrix}$ | $\begin{bmatrix}494.57 & 38.64\\38.64 & 19.44\end{bmatrix}$ |
| $i$ | 49 | 50 | 51 | 52 | 53 | 54 | 55 | 56 |
| $w_i$ | 0.0506 | -0.1374 | -0.0284 | 0.0551 | 0.0674 | -0.2179 | 0.1391 | -0.1446 |
| $\mu_i$ | $\begin{bmatrix}0.5631\\0.5057\end{bmatrix}$ | $\begin{bmatrix}0.0705\\0.8545\end{bmatrix}$ | $\begin{bmatrix}0.5632\\0.3883\end{bmatrix}$ | $\begin{bmatrix}0.8214\\0.5377\end{bmatrix}$ | $\begin{bmatrix}0.9980\\0.8566\end{bmatrix}$ | $\begin{bmatrix}0.5430\\0.6116\end{bmatrix}$ | $\begin{bmatrix}0.0873\\0.6527\end{bmatrix}$ | $\begin{bmatrix}0.0838\\0.7833\end{bmatrix}$ |
| $A_i$ | $\begin{bmatrix}82.82 & -8.06\\-8.06 & 130.56\end{bmatrix}$ | $\begin{bmatrix}882.14 & -101.63\\-101.63 & 1404.02\end{bmatrix}$ | $\begin{bmatrix}25.54 & 40.12\\40.12 & 358.82\end{bmatrix}$ | $\begin{bmatrix}9.7971 & -0.0002\\-0.0002 & 0.0001\end{bmatrix}$ | $\begin{bmatrix}65.11 & 22.87\\22.87 & 65.48\end{bmatrix}$ | $\begin{bmatrix}680.97 & -36.61\\-36.61 & 515.02\end{bmatrix}$ | $\begin{bmatrix}251.88 & -33.54\\-33.54 & 312.62\end{bmatrix}$ | $\begin{bmatrix}1783.98 & -309.81\\-309.81 & 3736.87\end{bmatrix}$ |
| $i$ | 57 | 58 | 59 | 60 | 61 | 62 | 63 | 64 |
| $w_i$ | -0.1343 | 0.3112 | 0.0659 | -0.0648 | -0.2009 | 0.0960 | 0.1069 | -0.0496 |
| $\mu_i$ | $\begin{bmatrix}0.8677\\0.6520\end{bmatrix}$ | $\begin{bmatrix}0.0778\\0.7830\end{bmatrix}$ | $\begin{bmatrix}0.7155\\0.8040\end{bmatrix}$ | $\begin{bmatrix}0.4064\\0.5952\end{bmatrix}$ | $\begin{bmatrix}0.0341\\0.9819\end{bmatrix}$ | $\begin{bmatrix}0.4872\\0.6399\end{bmatrix}$ | $\begin{bmatrix}0.4286\\0.4510\end{bmatrix}$ | $\begin{bmatrix}0.2876\\0.6934\end{bmatrix}$ |
| $A_i$ | $\begin{bmatrix}727.55 & 56.36\\56.36 & 59.44\end{bmatrix}$ | $\begin{bmatrix}900.18 & -269.49\\-269.49 & 2060.45\end{bmatrix}$ | $\begin{bmatrix}6.91 & 1.23\\1.23 & 63.22\end{bmatrix}$ | $\begin{bmatrix}234.36 & -61.10\\-61.10 & 90.22\end{bmatrix}$ | $\begin{bmatrix}906.36 & -416.93\\-416.93 & 324.42\end{bmatrix}$ | $\begin{bmatrix}0.0112 & -0.1367\\-0.1367 & 81.5282\end{bmatrix}$ | $\begin{bmatrix}23.3170 & -0.0070\\-0.0070 & 0.0471\end{bmatrix}$ | $\begin{bmatrix}216.42 & -112.35\\-112.35 & 571.51\end{bmatrix}$ |
| $i$ | 65 | 66 | 67 | 68 | 69 | 70 | 71 | 72 |
| $w_i$ | 0.0998 | -0.1082 | -0.0441 | 0.0874 | 0.0280 | -0.1154 | -0.0950 | 0.0719 |
| $\mu_i$ | $\begin{bmatrix}0.1977\\0.8542\end{bmatrix}$ | $\begin{bmatrix}0.4738\\0.7280\end{bmatrix}$ | $\begin{bmatrix}0.9961\\0.4953\end{bmatrix}$ | $\begin{bmatrix}0.9153\\0.7193\end{bmatrix}$ | $\begin{bmatrix}0.9498\\0.1216\end{bmatrix}$ | $\begin{bmatrix}0.1713\\0.3854\end{bmatrix}$ | $\begin{bmatrix}0.8191\\0.6194\end{bmatrix}$ | $\begin{bmatrix}0.6515\\0.8287\end{bmatrix}$ |
| $A_i$ | $\begin{bmatrix}177.13 & 17.27\\17.27 & 11.69\end{bmatrix}$ | $\begin{bmatrix}169.18 & -151.29\\-151.29 & 260.72\end{bmatrix}$ | $\begin{bmatrix}1468.18 & 61.49\\61.49 & 83.57\end{bmatrix}$ | $\begin{bmatrix}40.04 & -7.38\\-7.38 & 47.82\end{bmatrix}$ | $\begin{bmatrix}136.65 & -28.91\\-28.91 & 201.97\end{bmatrix}$ | $\begin{bmatrix}156.07 & 98.28\\98.28 & 206.71\end{bmatrix}$ | $\begin{bmatrix}836.84 & -51.35\\-51.35 & 123.18\end{bmatrix}$ | $\begin{bmatrix}88.17 & -11.71\\-11.71 & 105.81\end{bmatrix}$ |
| $i$ | 73 | 74 | 75 | 76 | 77 | 78 | 79 | 80 |
| $w_i$ | 0.0780 | 0.0866 | -0.1381 | -0.1555 | 0.1087 | -0.1090 | 0.0978 | -0.2910 |
| $\mu_i$ | $\begin{bmatrix}0.5903\\0.7526\end{bmatrix}$ | $\begin{bmatrix}0.7344\\1.0000\end{bmatrix}$ | $\begin{bmatrix}0.8508\\0.2861\end{bmatrix}$ | $\begin{bmatrix}0.3308\\0.4139\end{bmatrix}$ | $\begin{bmatrix}0.4927\\0.9676\end{bmatrix}$ | $\begin{bmatrix}0.5866\\0.7111\end{bmatrix}$ | $\begin{bmatrix}0.4398\\0.5432\end{bmatrix}$ | $\begin{bmatrix}0.0170\\0.8543\end{bmatrix}$ |
| $A_i$ | $\begin{bmatrix}444.30 & 149.46\\149.46 & 222.52\end{bmatrix}$ | $\begin{bmatrix}6.80 & -1.01\\-1.01 & 56.49\end{bmatrix}$ | $\begin{bmatrix}463.22 & 113.20\\113.20 & 132.97\end{bmatrix}$ | $\begin{bmatrix}58.79 & 13.04\\13.04 & 61.82\end{bmatrix}$ | $\begin{bmatrix}0.2603 & -0.5536\\-0.5536 & 53.1084\end{bmatrix}$ | $\begin{bmatrix}442.47 & 60.95\\60.95 & 195.94\end{bmatrix}$ | $\begin{bmatrix}23.8873 & 0.0545\\0.0545 & 0.3708\end{bmatrix}$ | $\begin{bmatrix}204.63 & -52.08\\-52.08 & 183.14\end{bmatrix}$ |

$i$ is the sequence number of Gaussian function. $w$ is the weight coefficient. $\mu$ denotes the Gaussian center. $A$ is a positive-definite symmetric precision matrix.

TABLE 4
REPRESENTATION RESULTS OF PERSONAL FACE IMAGES BY GMFACE WITH DIFFERENT QUANTITY OF GAUSSIAN COMPONENTS

| | m | original image | 40 | 50 | 60 | 70 | 80 | 90 |
|---|---|---|---|---|---|---|---|---|
| | Paramete Sizer | —— | 240 | 300 | 360 | 420 | 480 | 540 |
| | MSE | 0.0 | 0.000624 | 0.000515 | 0.000393 | 0.000251 | **0.000199** | 0.000182 |
| | PAE | 0.0 | 0.239427 | 0.231758 | 0.228463 | 0.214998 | **0.204020** | 0.175142 |
| Face Image 1 | Visual Effect | 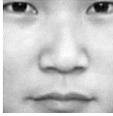 | 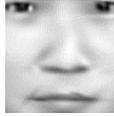 | 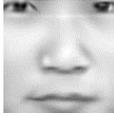 | 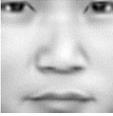 | 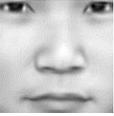 | 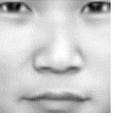 | 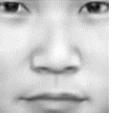 |
| | MSE | 0.0 | 0.000624 | 0.000515 | 0.000393 | 0.000251 | **0.000199** | 0.000182 |
| | PAE | 0.0 | 0.239427 | 0.231758 | 0.228463 | 0.214998 | **0.204020** | 0.175142 |
| Face Image 1 | Visual Effect | 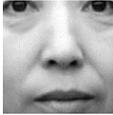 | 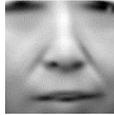 | 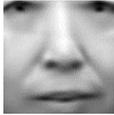 | 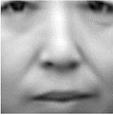 | 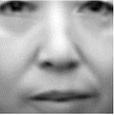 | 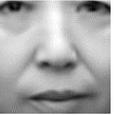 | 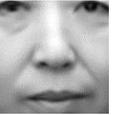 |
| Face Image 1 | MSE | 0.0 | 0.000624 | 0.000515 | 0.000393 | 0.000251 | **0.000199** | 0.000182 |
| | PAE | 0.0 | 0.239427 | 0.231758 | 0.228463 | 0.214998 | **0.204020** | 0.175142 |
| | Visual Effect | 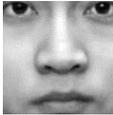 | 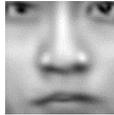 | 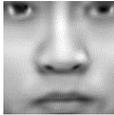 | 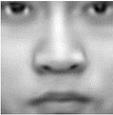 | 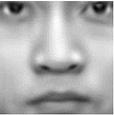 | 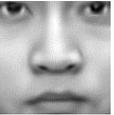 | 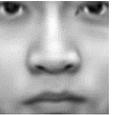 |

*m is the quantity of Gaussian components.*

common face model represented by GmFace provides almost no difference from the calculated average face.

To further dissect the face pattern modeling by GmFace, a number of significantly more important Gaussian components were selected from the aforementioned 80 Gaussian components to regenerate the model. As the peak value of each Gaussian unit is 1, Gaussian components were selected according to the absolute value of $w_i$, from the highest to the lowest. Here, k (k≤m) denotes the number of the chosen Gaussian components. The outputs of the GmNet with a portion of Gaussian components are shown in Fig. 9.

In the established GmFace denoted in Eq. (6), parameters $\mathbf{\mu}_i$ and $\mathbf{L}_i$ create different Gaussians spatial positions, sizes, and directions, and parameter $w_i$ determines the peak value besides the positive and negative of Gaussians. As illustrated in Fig. 9 (c), the 2D surface constructed by GmNet is not disordered, and is established according to its regulations in which the Gaussian components corresponding to larger absolute value of $w_i$ are used to portray important areas such as eyes, nose, and mouth. Among the total 80 Gaussian components, the approximate appearance of the face can be superposed by 40 Gaussian components with higher $w_i$. The remaining Gaussian components with lower $w_i$ act as local regulators for the generated face model. It can also be observed that, like real a human face, symmetry is reflected to a certain extent. Consequently, the validity of the proposed model GmFace for face modeling can be certified according to both parametric analysis and result visualization.

*3) Modeling Results of Personal Face*

The effectiveness of the proposed GmFace in personal face image representation was evaluated on CAS-PEAL-R1, and the representation results of three randomly selected samples were visualized. Under different parameter size, the reconstruction effect is as shown in Table 4.

The following conclusions can be drawn from the results in Table 4.

GmFace calculated by GmNet can provide good representation effect in visualization for personal modeling.

With the increase of the quantity of Gaussian components, the parameter size of the GmFace model grows gradually, the investigation index MSE is reduced, and the visual effect of face images reconstructed by GmFace becomes increasingly similar to the original images, especially the detail location.

When the quantity of Gaussian components reaches 80, the face reconstruction effect is significantly improved. Beyond this number, any further increase of Gaussian components provides little improvement in the reconstruction effect. Therefore, in the remaining experiments, the quantity of Gaussian components m is set to 80.

*4) Image Transformation Verification Results*

Using "Face Image 1" in Table 4 as an example, based on the solved GmFace model with 80 Gaussian components, the image translation, scaling, and rotation are realized using Eq. (15), (16), and (17).

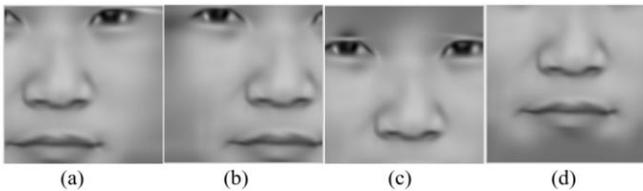

Fig. 10. Image translation processing result using GmFace. (a) Move left. (b) Move right. (c) Move up. (d) Move down.

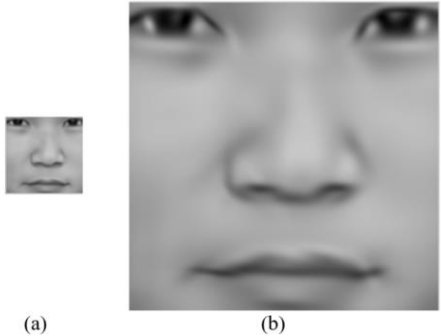

Fig. 11. Image scaling processing result through GmFace. (a) Scaling down. (b) Scaling up.

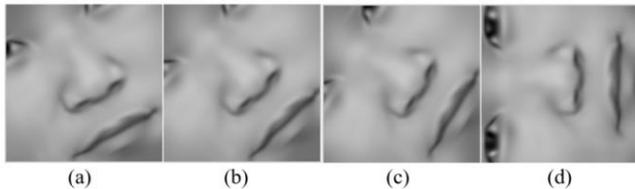

Fig. 12. Image rotation processing result through GmFace. (A) Rotate 30°. (B) Rotate 45°. (B) Rotate 60°. (C) Rotate 90°.

For image translation processing, the translational vector $\bar{\mathbf{x}} = \begin{bmatrix} 0 \\ 0.2 \end{bmatrix}, \begin{bmatrix} 0 \\ -0.2 \end{bmatrix}, \begin{bmatrix} 0.2 \\ 0 \end{bmatrix}, \begin{bmatrix} -0.2 \\ 0 \end{bmatrix}$, which represents that the translation direction is left, right, up, and down, respectively. The image translation results are illustrated in Fig. 10.

For image scaling processing, the scaling factor $k = 2, \frac{1}{2}$, which denotes scaling down two times and scaling up two times, respectively. The image scaling results are illustrated in Fig. 11.

For image rotation processing, the rotation angle $\theta = 30°, 45°, 60°, 90°$ and the center of rotation $\bar{\mathbf{x}} = \begin{bmatrix} 0.5 \\ 0.5 \end{bmatrix}$, which means that the image will rotate at different angles around the central point. The image rotation results are provided in Fig. 12.

According to the results of the image transformation above, it can be observed that the three operations of translation, scaling, and rotation are well realized by GmFace model. Compared to traditional methods of pixel operations, the proposed GmFace is a continuous transformation function of gray value correlated with position. In the proposed method, the face image can be transformed by parameter adjustment without complicated interpolation transformation. Once the GmFace model of a face is constructed, the transformation becomes very simple. Moreover, as illustrated visually in the experimental results, there is an additional smooth effect.

## V. CONCLUSION

Contrasting from traditional methods based on hand-crafted features and data-driven deep neural network learning methods, the explicit mathematical representation of human face was the focus of this work. The face representation mathematical model GmFace was proposed in this paper along with the neural network GmNet which was constructed to solve the model parameters. Furthermore, with this representation model, face image transformation can be realized mathematically through simple parameter computation.

According to Occam's razor, the simplest solution is most likely the right one. Perhaps, GmFace is not the simplest model for face representation, but it has taken the first step towards this goal. In the future, one of the further works to do is to analyze the characteristics of GmFace in depth and investigate facial vision applications based on it. The other study is to explore the simplest face model by replacing the multi-Gaussian function in GmFace with other elementary functions, such as exponential, trigonometric, logarithmic or composite functions.

Here we propose a new term – *deep modeling*, which refer to the process of developing a mathematical model based on deep learning to gain deep insight into a complex black-box system. This work is also a tentative study on deep modeling.

ACKNOWLEDGMENT

This work was supported by the National Science Foundation of China under Grants 61901436 and 61572458.

The corresponding author is Weijun Li.

REFERENCES

[1] I. Oruc, B. Balas, and M. S. Landy, "Face Perception: A Brief Journey through Recent Discoveries and Current Directions," *Vision Res.*, vol.157, pp.1-9, Apr. 2019, doi: 10.1016/j.visres. 2019.06.005.

[2] G. Guo and N. Zhang, "A Survey on Deep Learning Based Face Recognition," *Comput. Vis. Image Und.*, vol.189, pp.1-37, Dec. 2019, doi: 10.1016/j.cviu.2019.102805.

[3] A. Timo, H. Abdenour, and P. Matti, "Face Recognition with Local Binary Patterns," in *Proc. Eur. Conf. Comput. Vis. (ECCV'04)*, Prague, CZE, pp.469-481, 2004, doi: 10.1007/978-3-540-24670-1_36.

[4] S. Shan, W. Gao, Y. Chang, B. Cao, and P. Yang, "Review the Strength of Gabor Features for Face Recognition from the Angle of its Robustness to Misalignment," in *Proc. Int. Conf. Pattern Recognition (ICPR'04)*, Cambridge, UK, pp.338-341, 2004, doi:10.1109/ICPR.2004.1334121

[5] T. Lindeberg, "Scale Invariant Feature Transform," *Scholarpedia*, Chapter 7(5), pp.10491, 2012, doi: 10.4249/scholarpedia.10491.

[6] S. Chang, X. Ding, and C. Fang. "Histogram of the Oriented Gradient for Face Recognition." *Tsinghua Sci. Technol.*, vol.16, no.2, pp.216-224, 2011, doi: 10.1016/s10070214(11)70032-3.

[7] M. Turk and A. Pentland, "Eigenfaces for Recognition," *J Cognitive Neurosci.*, vol.3, no.1, pp.71-86, Jan. 1991, doi: 10.1162/jocn.1991.3.1.71.

[8] P. N. Belhumeur, J. P. Hespanha, and D. J. Kriegman, "Eigenfaces vs. Fisherfaces: Recognition Using Class Specific Linear Projection," in *Proc. Eur. Conf. Comput. Vis. (ECCV), Cambridge*, UK, 1996, pp.43-58, doi: 10.1007/BFb0015522.

[9] M. Wang and W. Deng, "Deep Face Recognition: A Survey," Feb. arXiv preprint, available at https://arxiv.org/abs/1804.06655, 2019.

[10] Y. Sun, X. Wang, and X. Tang, "Deep Learning Face Representation from

Predicting 10,000 Classes," in *Proc. IEEE Conf. Comput. Vis. Pattern Recognit. (CVPR'14)*, Columbus, OH, USA, 2014, pp.1891-1898, doi: 10.1109/CVPR.2014.244.

[11] Y. Sun, X. Wang, and X. Tang, "Deeply Learned Face Representation are Sparse, Selective, and Robust," arXiv preprint, available at https://arxiv.org/abs/1412.1265, Dec. 2014.

[12] A. Krizhevsky, I. Sutskever, and G. E. Hinton, "ImageNet Classification with Deep Convolutional Neural Networks," *Commun. ACM*, vol.60, no.6, pp.84-90, Jun. 2017, doi: 10.1145/3065386.

[13] Y. Taigman, M. Yang, M. Ranzato, and L. Wolf, "DeepFace: Closing the Gap to Human-Level Performance in Face Verification," in *Proc. IEEE Conf. Comput. Vis. Pattern Recognit. (CVPR'14)*, Columbus, OH, USA, pp.1701-1708, 2014, doi: 10.1109/CVPR.2014.220.

[14] F. Schroff, D. Kalenichenko, and J. Philbin, "FaceNet: A Unified Embedding for Face Recognition and Clustering," in *Proc. IEEE Conf. Comput. Vis. Pattern Recognit. (CVPR'15)*, Boston, MA, USA, pp. 815-823, 2015, doi: 10.1109/CVPR.2015.7298682.

[15] W. Liu, Y. Wen, Z. Yu, M. Li, B. Raj, and L. Song, "SphereFace: Deep Hypersphere Embedding for Face Recognition," in *Proc. IEEE Conf. Comput. Vis. Pattern Recognit. (CVPR'17)*, Honolulu, HI, USA, pp. 6738-6746, 2017, doi: 10.1109/CVPR.2017.713.

[16] Y. LeCun, L. Bottou, Y. Bengio, and P. Haffner, "Gradient-Based Learning Applied to Document Recognition," *P. IEEE*, vol.86, no.11, pp.2278–2324, Dec. 1998, doi: 10.1109/5.726791.

[17] K. Simonyan and A. Zisserman, "Very Deep Convolutional Networks for Large-Scale Image Recognition," arXiv preprint, available at https://arxiv.org/abs/1409.1556, Sep. 2014.

[18] C. Szegedy, L. Wei, Y. Jia, P. Sermanet, S. Reed, D. Anguelov, D. Erhan, V. Vanhoucke, and A, Rabinovich, "Going Deeper with Convolutions," in *Proc. IEEE Conf. Comput. Vis. Pattern Recognit. (CVPR'15)*, Boston, MA, USA, pp.1-9, 2015, doi: 10.1109/CVPR.2015.7298594.

[19] K. He, X. Zhang, S. Ren, and J. Sun, "Deep Residual Learning for Image Recognition," in *Proc. IEEE Conf. Comput. Vis. Pattern Recognit. (CVPR'16)*, Las Vegas, NV, USA, pp.770-778, 2016, doi: 10.1109/CVPR.2016.90.

[20] T. Clark and E. Nyberg, "Creating the Black Box: A Primer on Convolutional Neural Network Use in Image Interpretation," *Current Problems in Diagnostic Radiology*, Jul. 2019, doi: 10.1067/j.cpradiol.2019.07.004.

[21] X. Li, W. Zhang, and Q. Ding, "Understanding and Improving Deep Learning-based Rolling Bearing Fault Diagnosis with Attention Mechanism," *Signal Process.*, vol.161, pp.136-154, Aug. 2019, doi: 10.1016/j.sigpro.2019.03.019.

[22] J. D. Olden and D. A. Jackson, "Illuminating the "Black Box": A Randomization Approach for Understanding Variable Contributions in Artificial Neural Networks," *Ecol. Model.*, vol.154, no.1-2, pp.135-150, Aug. 2002, doi: 10.1016/S0304-3800(02)00064-9.

[23] I. M. Revina and W. R. S. Emmanuel, "A Survey on Human Face Expression Recognition Techniques," *Journal of King Saud University-Computer and Information Sciences*, Sep. 2018, doi:10.1016/j.jksuci.2018.09.002.

[24] M. MacLeod and N. J. Nersessian, "Mesoscopic Modeling as a Cognitive Strategy for Handling Complex Biological Systems," *Studies in History and Philosophy of Biol. & Biomed. Sci.*, Aug. 2019, doi: 10.1016/j.shpsc.2019.101201.

[25] P. Kohl, E.J. Crampin, T.A. Quinn, and D. Noble. "Systems Biology: An Approach," *Clin. Pharmacol. Ther.*, vol.88, no.1, pp.25-33, Jul. 2010, doi: 10.1038/clpt.2010.92.

[26] T. Melham, "Modelling, Abstraction, and Computation in Systems Biology: A View from Computer Science," *Prog. Biophys. Mol. Bio.*, vol.111, no.2-3, pp.129-136, Apr. 2013, doi: 10.1016/j.pbiomolbio.2012.08.015.

[27] "How to Integrate Gaussian Functions," wikiHow, Oct. 1, Available at https://www.wikihow.com/Integrate-Gaussian-Functions, 2017.

[28] L. Wang, L. Xu, S. Feng, Q. H. Meng, and K. Wang, "Multi-Gaussian Fitting for Pulse Waveform Using Weighted Least Squares and Multi-criteria Decision Making Method," *Comput. Biol. Med.*, vol.43, no.11, pp.1661-1672, Nov. 2013, doi: 10.1016/j.compbiomed.2013.08.004.

[29] U. Rubins, "Finger and Ear Photoplethysmogram Waveform Analysis by Fitting with Gaussians," *Med. Biol. Eng. Comput.*, vol.46, no.12, pp.1271-1276, 2008, doi: 10.1007/s11517-008-0406-z.

[30] M. C Baruch, D. ER Warburton, S. SD Bredin, A. Cote, D. W Gerdt, and C. M Adkins, "Pulse Decomposition Analysis of the Digital Arterial Pulse During Hemorrhage Simulation, " *Nonlinear Biomed. Phys.*, vol.5, no.1, 2011, doi: 10.1186/1753-4631-5-1.

[31] C. Liu, D. Zheng, A. Murray, and C. Liu, "Modeling Carotid and Radial Artery Pulse Pressure Waveforms by Curve Fitting with Gaussian Functions," *Biomed Signal Process Control*, vol.8, no.5, pp.449-454, Sep. 2013, doi: 10.1016/j.bspc.2013.01.003.

[32] R. Couceiro, P. Carvalho, R.P. Paiva, J. Henriques, I. Quintal, M. Antunes, J. Muehlsteff, C. Eickholt, C. Brinkmeyer, M. Kelm, and C. Meyer, "Assessment of Cardiovascular Function from Multi-Gaussian Fitting of a Finger Photoplethysmogram," *Physiol. Meas.*, vol.36, no.9, pp.1801-1825, Sep. 2015, doi: 10.1088/0967-3334/36/9/1801

[33] H. Zhang, L. Xu, S. Song, and R. Jiao, "A Fast Method for X-ray Pulsar Signal Simulation," *Acta Astronaut.*, vol.98, pp.189-200, May-Jun. 2014, doi: 10.1016/j.actaastro.2014.01.030.

[34] X. Xu and X. Wu, "Mean Pulse Analysis and Spectral Character Study of Pulsar PSR B2111+46," *Sci. China (Ser. G.)*, vol.46, no.1, pp.104-112, 2003, doi: 10.1360/03yg9015.

[35] M. Kramer, R. Wielebinski, A. Jessner, J. A. Gil, and J. H. Seiradakis, "Geometrical Analysis of Average Pulsar Profiles Using Multi-component Gaussian FITS at Several Frequencies. I. Method and analysis," *Astron. Astrophys.*, vol.107, pp.515-526, 1994.

[36] R.Y. Xu and M. Kemp, "Fitting Multiple Connected Ellipses to an Image Silhouette Hierarchically," *IEEE T. Image Process.*, vol.19, no.7, pp.1673-1682, Jul. 2010, doi: 10.1109/TIP.2010.2045071.

[37] A. Goshtasby and W. D. O'Neill, "Surface Fitting to Scattered Data by a Sum of Gaussians," *Comput. Aided. Geom. D.*, vol.10, no.2, pp.143-156, 1993, doi: 10.1016/0167-8396(93)90017-w.

[38] R. Wilson, "Multiresolution Gaussian Mixture Models: Theory and Applications," *Department of Computer Science*, Univ. Warwick, Coventry, UK, Feb. 28, 2000.

[39] S. Zhao, "Pixel-level Occlusion Detection Based on Sparse Representation for Face Recognition," *Optik*, vol. 168, pp.920-930, Sep. 2018, doi: 10.1016/j.ijleo.2018.05.013.

[40] J.Shen, X. Zuo, J. Li, W.Yang, and H. Ling, "A Novel Pixel Neighborhood Differential Statistic Feature for Pedestrian and Face Detection," *Pattern Recogn.*, vol. 63, pp.127-138, Mar. 2017, doi: 10.1016/j.patcog.2016.09.010.

[41] A. Asthana, S. Zafeiriou, G. Tzimiropoulos, S. Cheng, and M. Pantic, "From Pixels to Response Maps: Discriminative Image Filtering for Face Alignment in the Wild," *IEEE T. Pattern. Anal.*, vol. 37, no. 6, pp. 1312-1320, Oct. 2014, doi: 10.1109/TPAMI.2014.2362142.

[42] S. Choi, C. Choi, G. Jeong, and N. Kwak, "Pixel Selection Based on Discriminant Features with Application to Face Recognition," *Pattern Recogn. Lett.*, vol. 33, no. 9, pp.1083-1092, Jul. 2012, doi: 10.1016/j.patrec.2012.01.005.

[43] G. Tzimiropoulos, S. Zafeiriou, and M. Pantic, "Subspace Learning from Image Gradient Orientations," *IEEE T. Pattern Anal.*, vol. 34, no. 12,

pp.2454-2466, Jan. 2012, doi: 10.1109/TPAMI.2012.40.

[44] M. S. Nixon and A. S. Aguado, "Chapter 3 - Basic Image Processing Operations," in *Feature Extraction & Image Processing for Computer Vision*, 3th ed. London, UK: Elsevier, ch.3, pp.105, available at https://www.sciencedirect.com/science/article/pii/B9780123965493000033, 2012.

[45] Z. Dostál, T. Kozubek, A. Markopoulos, and M. Menšíka, "Cholesky Decomposition of a Positive Semidefinite Matrix with Known Kernel," *Appl. Math. Ccmput.*, vol.217, no.13, pp.6067-6077, Mar. 2011, doi: 10.1016/j.amc.2010.12.069

[46] D. E. Rumelhart, G. E. Hinton, and R. J. Williams, "Learning Representations by Back Propagating Errors," *Nature*, vol.323, no. 6088, pp. 533-536, Oct. 1986, doi: 10.1038/323533a0.

[47] D. P. Kingma and J. Ba, "Adam: a Method for Stochastic Pptimization," arXiv preprint, available at https://arxiv.org/abs/1412.6980, Jan. 2017.

[48] S. Ruder, "An Overview of Gradient Descent Optimization Algorithms," arXiv preprint, available at https://arxiv.org/abs/1609.04747, Jun. 2017.

[49] W. Gao, B. Cao, and S. Shan, "The CAS- PEAL Large-scale Chinese Face Database and Evaluation Protocols," *Joint Research & Development Laboratory*, CAS, Beijing, China, No.JDL_TR_04_FR_001, 2004.

[50] P. Adam, S. Gross, S. Chintala, G. Chanan, E. Yang, Z. DeVito, Z. Lin, A. Desmaison, L. Antiga, and A. Lerer, "Automatic Differentiation in PyTorch," In *NIPS 2017 Workshop Autodiff Program Chairs*, Oct. 2017.